\documentclass[10pt,twocolumn,letterpaper]{article}

\usepackage[pagenumbers]{cvpr} 

\usepackage{graphicx}
\usepackage{amsmath}
\usepackage{amssymb}
\usepackage{booktabs}

\usepackage{url}
\usepackage{wrapfig} 
\usepackage{multirow}
\usepackage{booktabs}
\usepackage{siunitx}
\graphicspath{{figure/}}

\usepackage{appendix}
\usepackage[american]{babel} 
\usepackage{pifont}
\usepackage[cmyk]{xcolor}
\usepackage[accsupp]{axessibility}

\newcommand{\PractiseNoSpace}{\textsc{Practise}} 
\newcommand{\Practise}{\textsc{Practise} } 
\newcommand{\PractiseEmph}{\emph{\textsc{Practise}} } 

\makeatletter
\newif\if@restonecol
\makeatother

\usepackage[ruled,vlined]{algorithm2e}
\usepackage{algpseudocode}

\usepackage[pagebackref,breaklinks,colorlinks]{hyperref}

\def\cvprPaperID{6568}
\def\confName{CVPR}
\def\confYear{2023}

\begin{document}

\title{Practical Network Acceleration with Tiny Sets}

\author{
Guo-Hua Wang\quad\quad Jianxin Wu\thanks{J. Wu is the corresponding author. This research was partly supported by the National Natural Science Foundation of China under Grant 62276123 and Grant 61921006.}\\ 
State Key Laboratory for Novel Software Technology\\ 
Nanjing University, Nanjing, China\\
{\tt\small wangguohua@lamda.nju.edu.cn}, {\tt\small wujx2001@nju.edu.cn}
}

\maketitle

\begin{abstract}
Due to data privacy issues, accelerating networks with tiny training sets has become a critical need in practice. Previous methods mainly adopt filter-level pruning to accelerate networks with scarce training samples. In this paper, we reveal that dropping blocks is a fundamentally superior approach in this scenario. It enjoys a higher acceleration ratio and results in a better latency-accuracy performance under the few-shot setting. To choose which blocks to drop, we propose a new concept namely recoverability to measure the difficulty of recovering the compressed network. Our recoverability is efficient and effective for choosing which blocks to drop. Finally, we propose an algorithm named \PractiseEmph to accelerate networks using only tiny sets of training images. \PractiseEmph outperforms previous methods by a significant margin. For 22\% latency reduction, \PractiseEmph surpasses previous methods by on average 7\% on ImageNet-1k. It also enjoys high generalization ability, working well under data-free or out-of-domain data settings, too. Our code is at \url{https://github.com/DoctorKey/Practise}. 
\end{abstract}

\section{Introduction}
\label{sec:intro}

In recent years, convolutional neural networks (CNNs) have achieved remarkable success, but they suffer from high computational costs. To accelerate the networks, many network compression methods have been proposed, such as network pruning~\cite{L1_norm,thinet,slimming,JointPruning}, network decoupling~\cite{DW_PW,LRD} and network quantization~\cite{quantization,bannerscalable}. However, most previous methods rely on \emph{the original training set} (i.e., all the training data) to recover the model's accuracy. But, to preserve data privacy and/or to achieve fast deployment, only scarce training data may be available in many scenarios. 

For example, a customer often asks the algorithmic provider to accelerate their CNN models, but due to privacy concerns, the whole training data \emph{cannot} be available. Only the raw uncompressed model and a few training examples are presented to the algorithmic provider. In some extreme cases, not even a single data point is to be provided. The algorithmic engineers need to synthesize images or collect some out-of-domain training images by themselves. Hence, to learn or tune a deep learning model with only very few samples is emerging as a critical problem to be solved.

\begin{figure}
	\centering
	\includegraphics[width=0.75\linewidth]{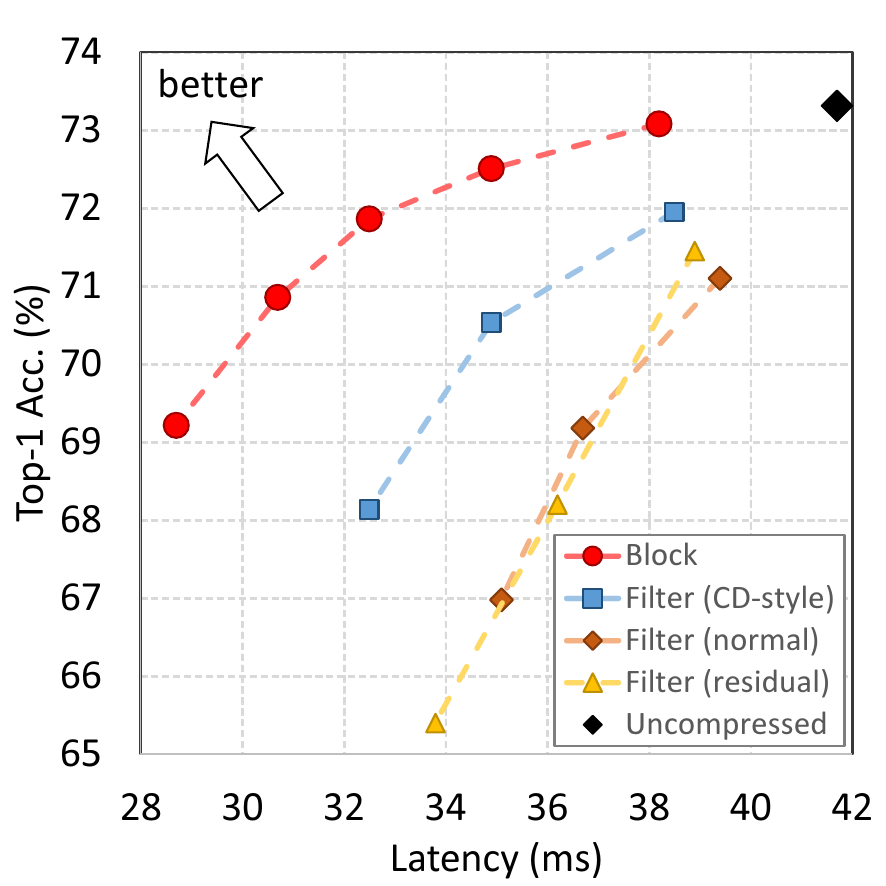}
	\caption{Comparison of different compression schemes with only 500 training images. We propose dropping blocks for few-shot network acceleration. Our method (`Block') outperforms previous methods dominantly for the latency-accuracy tradeoff. The ResNet-34 model was compressed on ImageNet-1k and all latencies were tested on an NVIDIA TITAN Xp GPU.}
	\label{fig:top1-latency}
\end{figure}

In this few-shot compression scenario, most previous works~\cite{FSKD,CD,MiR} adopt filter-level pruning. However, it \emph{cannot achieve a high acceleration ratio} on real-world computing devices (e.g., on GPUs). To make compressed models indeed run faster than the uncompressed models, lots of FLOPs (number of floating point operations) are required to be reduced by filter-level pruning. And without the whole training dataset, it is difficult to recover the compressed model's accuracy. Hence, previous few-shot compression methods often exhibit a poor latency (wall-clock timing) vs. accuracy tradeoff. 

\begin{table}
	\centering
  \small
	\begin{tabular}{ccccc}
		\toprule
    KD~\cite{KD} & FSKD~\cite{FSKD}	& CD~\cite{CD} & MiR~\cite{MiR} & BP (blocks)	\\
		\midrule
    44.5 & 45.3 & 56.2 & 64.1 & \textbf{66.5} \\
		\bottomrule
	\end{tabular}
  \caption{Top-1 validation accuracy (\%) on ImageNet-1k for different compression schemes. ResNet-34 was accelerated by reducing $16\%$ latency with 50 training images. Previous methods prune filters with the `normal' style. For the block-level pruning, we simply remove the first $k$ blocks and finetune the pruned network by back propagation, i.e., `BP (blocks)' in this table. }
  \label{tab:drop-block-BP}
\end{table}

In this paper, we advocate that we need to focus on latency-accuracy rather than FLOPs-accuracy, and reveal that block-level pruning is fundamentally superior in the few-shot compression scenario. Compared to pruning filters, dropping blocks enjoys a higher acceleration ratio. Therefore it can keep more capacity from the original model and its accuracy is easier to be recovered by a tiny training set under the same latency when compared with filter pruning. Fig.~\ref{fig:top1-latency} shows dropping blocks dominantly outperforms previous compression schemes for the latency-accuracy tradeoff. Table~\ref{tab:drop-block-BP} further reports that \emph{an embarrassingly simple dropping block baseline (i.e., finetune without any other processing) has already surpassed existing methods which use complicated techniques}. The baseline, `BP (blocks)', simply removes the first few blocks and finetune the pruned network with the cross-entropy loss.

To further improve block pruning, we study the strategy for choosing which blocks to drop, especially when only scarce training samples are available. Several criteria~\cite{DBP,e_ResNet,CURL} have been proposed for pruning blocks on the whole dataset. However, some~\cite{DBP,e_ResNet} require a large amount of data for choosing, whereas others~\cite{CURL} only evaluate the output difference before/after block removal. In this paper, we notice that although dropping some blocks significantly changes the feature maps, they are easily recovered by end-to-end finetuning even with a tiny training set. So simply measuring the difference between pruned/original networks is not valid. To deal with these problems, a new concept namely \emph{recoverability} is proposed in this paper for better indicating blocks to drop. And we propose a method to compute it efficiently, with only a few training images. At last, our recoverability is surprisingly consistent with the accuracy of the finetuned network. 

Finally, we propose \PractiseNoSpace, namely \underline{Pr}actical network \underline{ac}celeration with \underline{ti}ny \underline{se}ts of images, to effectively accelerate a network with scarce data. \Practise significantly outperforms previous few-shot pruning methods. For $22.1\%$ latency reduction, \Practise surpasses the previous state-of-the-art (SOTA) method on average by \emph{7.0\%} (percentage points, not relative improvement) Top-1 accuracy on ImageNet-1k. It is also robust and enjoys high generalization ability which can be used on synthesized/out-of-domain images. Our contributions are:

$\bullet$ We argue that the FLOPs-accuracy tradeoff is a misleading metric for few-shot compression, and advocate that the latency-accuracy tradeoff (which measures real runtime on devices) is more crucial in practice. For the first time, we find that in terms of latency vs. accuracy, block pruning is an embarrassingly simple but powerful method---dropping blocks with simple finetuning has already surpassed previous methods (cf. Table~\ref{tab:drop-block-BP}). Note that although dropping blocks is previously known, we are \emph{the first to reveal its great potential in few-shot compression}, which is both a surprising and an important finding.

$\bullet$ To further boost the latency-accuracy performance of block pruning, we study the optimal strategy to drop blocks. A new concept \emph{recoverability} is proposed to measure the difficulty of recovering each block, and in determining the priority to drop blocks. Then, we propose \PractiseNoSpace, an algorithm for accelerating networks with tiny sets of images. 

$\bullet$ Extensive experiments demonstrate the extraordinary performance of our \PractiseNoSpace. In both the few-shot and even the extreme data-free scenario, \Practise improves results by a significant margin. It is versatile and widely applicable for different network architectures, too.

\section{Related Works} 

\textbf{Filter-level pruning} accelerates networks by removing filters in convolutional layers. Different criteria for choosing filters have been proposed~\cite{he2019filter,molchanov2019importance,L1_norm,slimming,thinet,CURL}, along with different training strategies~\cite{liu2018rethinking,ResRep,li2022revisiting,liu2021discrimination}. However, most these methods rely on the whole training data to train the network. When facing data privacy issues, these filter-level pruning methods suffer from poor latency-accuracy performance with tiny training sets~\cite{FSKD,CD}. In this paper, we argue that the main drawback of filter pruning is its low acceleration ratio. It requires pruning lots of parameters and FLOPs to reduce latency. That results in a large capacity gap between the pruned and the original networks. So it is challenging to recover the pruned network accuracy on only a tiny training set. Instead, we advocate block-level pruning for few-shot compression. Dropping blocks enjoys a higher acceleration ratio. We claim that it is a superior way to accelerate networks with only tiny training sets.

\textbf{Block-level pruning} removes the whole block (\eg, a residual block) in a network. Some works have been proposed for the whole training data case, but rarely studied in the few-shot scenario. BlockDrop~\cite{wu2018blockdrop} introduces a reinforcement learning approach to derive instance-specific inference paths in ResNets. DBP~\cite{DBP} proposes using linear probing to evaluate the accuracy of each block's features, and dropping blocks with low accuracy. $\epsilon$-ResNet adds a sparsity-promoting function to discard the block if all responses of this block are less than a threshold $\epsilon$. Both DBP and $\epsilon$-ResNet require a large dataset for training and testing. CURL~\cite{CURL} uses a proxy dataset to evaluate the KL-divergence change before/after block removal. However, it neglects the finetuning process. Actually, we care more about the accuracy of the pruned network \emph{after finetuning}. But there are no existing criteria to measure it well.

In this paper, we propose a new concept named \emph{recoverability} to evaluate if the network pruned by dropping blocks can recover the accuracy well. Our method for computing recoverability is efficient that only requires a few training samples. And it is effective to predict the accuracy of the finetuned network. Based on it, we propose \Practise, an algorithm for practical network acceleration with tiny sets. Our \Practise outperforms previous few-shot compression methods by a significant margin.

\textbf{Data Limited Knowledge Distillation} aims at training a student network by a pretrained teacher with limited original training data. Few-Shot Knowledge Distillation (FSKD)~\cite{FSKD} inserts $1\times 1$ conv. after the pruned conv. layer and trains each layer by making the pruned network's feature maps mimic the original network's. The layer-wisely training can obtain more supervised signals from the teacher, but easily results in error accumulation. CD~\cite{CD} proposes cross distillation to reduce the layer-wisely accumulated errors. MiR~\cite{MiR} proposes a mimicking then replacing framework to optimize the pruned network holistically. For a more extreme case, not even a single original training sample is available, Data Free Knowledge Distillation (DFKD) was proposed in~\cite{DI}. The core of DFKD is to synthesize alternatives of the original training data. Due to the promising results, more and more studies try to improve it, such as accelerating the synthesis process~\cite{fang2022up} and enhancing the performance by multi-teacher~\cite{MixMix}. Pruning and quantization~\cite{liu2021zero,zhang2021diversifying,cai2020zeroq} are two main applications of DFKD.

However, most FSKD and DFKD methods adopt filter-level pruning to compress networks and result in poor latency-accuracy performance. We claim that dropping blocks is a more data-efficient acceleration scheme. Our \Practise outperforms previous FSKD works significantly. It is also robust to work on synthesized/out-of-domain images and improves the accuracy in the data-free scenario by a large margin. 

\section{The Proposed Method}

First, we analyze the benefits of dropping blocks (\eg, dropping residual blocks in ResNet). Compared with previous few-shot compression methods, dropping blocks enjoys a high acceleration ratio that achieves superior latency-accuracy performance with the tiny training set. Second, we propose a new concept named recoverability for choosing which blocks to drop. Different from previous criteria, the recoverability measures the hardness of finetuning a pruned network, which is closely and more directly related to the model's accuracy. The recoverability is also efficient to compute, which suits the few-shot scenario well. Based on the recoverability, we propose \PractiseNoSpace, an algorithm for practical network acceleration with tiny training sets. Our \Practise does not require the label of the training set, and it is even able to work without using any image from the original training dataset.

\subsection{The motivation to drop blocks}

\begin{figure}
	\centering
	\includegraphics[width=0.75\linewidth]{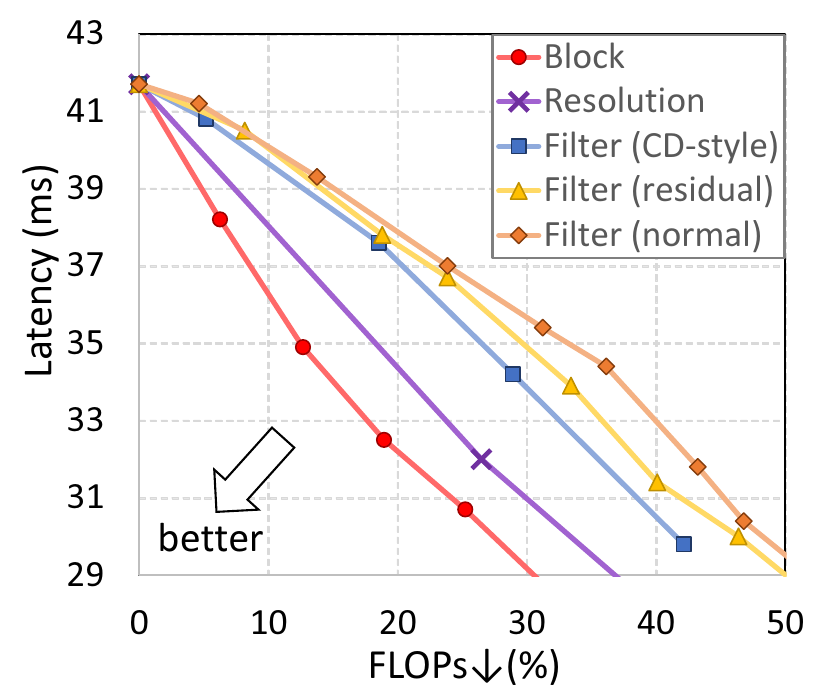}
	\caption{The relationship between latency and FLOPs reduction. Dropping blocks enjoys the highest acceleration ratio compared with other compression schemes.}
	\label{fig:latency-flops}
\end{figure}

For accelerating neural networks in real-world applications, latency and accuracy are the two most important metrics. Lower latency means the model runs faster on devices. To accelerate the model with \emph{tiny} training sets, many compression schemes have been proposed. Fig.~\ref{fig:latency-flops} compares the acceleration ratios of different schemes. FSKD~\cite{FSKD} prunes filters within residual blocks according to the $L_1$ norm (namely `normal'). CD~\cite{CD} proposes pruning conv. layers only in shallow layers, and keeps deeper conv. layers unchanged (namely `CD-style'). MiR~\cite{MiR} trims the residual connection (namely `residual'). All these pruning schemes suffer from inefficient acceleration ratios. As shown in Fig.~\ref{fig:latency-flops}, with about $30\%$ FLOPs reduction, compressed models achieve only 16.1\% latency reduction (41.7$\rightarrow$35 ms). Simply resizing the input image's resolution (namely `Resolution') achieves better acceleration. At last, dropping blocks outperforms all these methods. To achieve 35 ms latency, it only needs to reduce 12.7\% FLOPs, significantly less than $30\%$ in pruning filters.

Obviously, dropping blocks is more effective for model acceleration than pruning filters, but it is neglected in few-shot compression. One possible reason is that pruning filters has achieved extraordinary performance with the \emph{whole} training dataset~\cite{ResRep}. However, when only a tiny training set is available, finetuning the pruned model suffers from overfitting and unstable problems, especially for large FLOPs reductions~\cite{FSKD,MiR}. For the same latency reduction, dropping blocks keeps more parameters and capacity from the original model. Therefore, it requires less data for finetuning and achieves a superior latency-accuracy tradeoff with the \emph{tiny} training set (cf. Fig.~\ref{fig:top1-latency}). We have demonstrated that even a naive dropping blocks method has already outperformed most existing methods.

\begin{figure}
	\centering
	\includegraphics[width=1\linewidth]{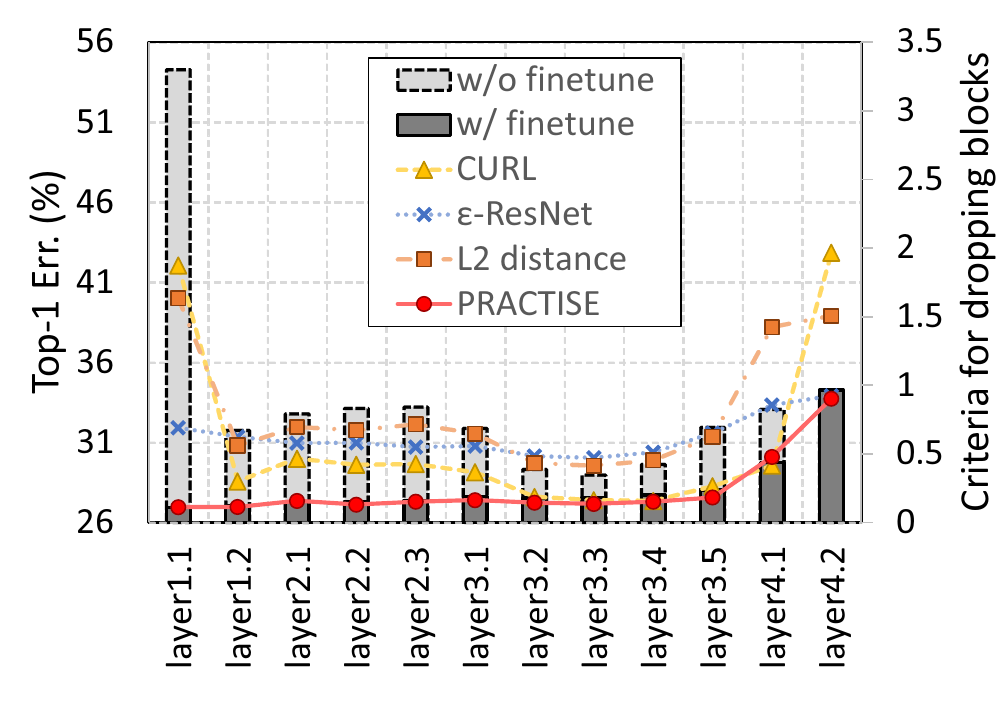}
	\caption{Illustration of the recoverability of each block and different criteria for dropping blocks. ResNet-34 has 12 blocks that can be dropped from `layer1.1' to `layer4.2'. We drop each block and evaluate the Top-1 error with/without finetuning, respectively. Scores of different criteria are computed for each block and presented in this figure, too. Note that the Top-1 error (\%) is evaluated on the ImageNet-1k validation set (50000 images), while finetuning and evaluating criteria take only 500 training images. \Practise predicts the finetuned network's error almost perfectly.}
	\label{fig:metric}
\end{figure}

\begin{figure*}
	\centering
	\includegraphics[width=0.78\linewidth]{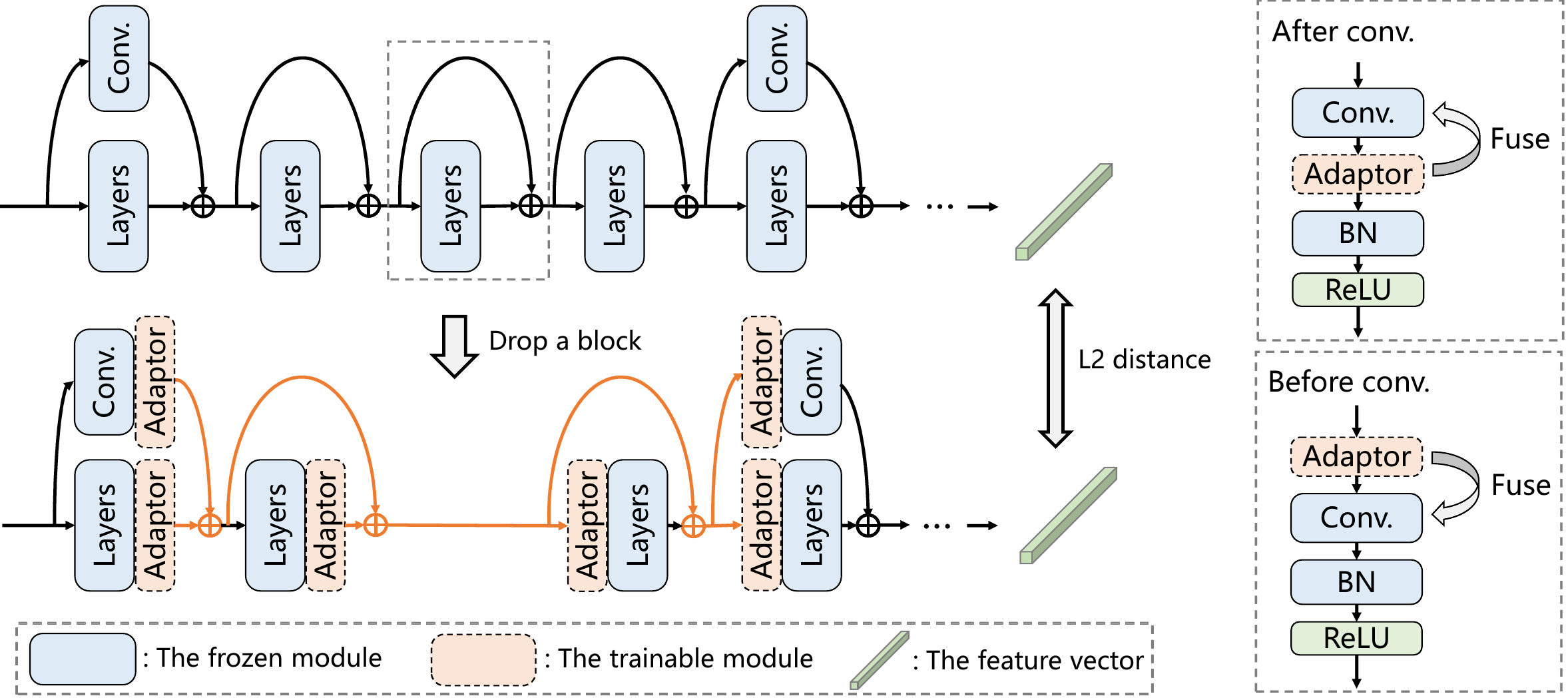}
	\caption{Illustration of our method for determining which block to drop. One block is dropped, and adaptors are inserted around the dropped position to alleviate the loss from dropping this block. Note that adaptors are \emph{not} added to blocks before the previous or after the next downsample operation. Then, the L2 distance of pruned/original networks' features is computed as the \emph{recoverability} of this dropped block. Finally, we drop several blocks that are easily recovered to obtain the pruned network. }
	\label{fig:practise}
\end{figure*}

\subsection{The recoverability of the pruned model}

To further improve dropping blocks, we study how to choose blocks to drop. Different from previous works~\cite{DBP,e_ResNet,CURL} that neglect the finetuning process, we argue that an effective metric should be consistent with the finetuned (\ie, recovered) accuracy. To this end, we propose a new concept namely \emph{recoverability}. Recoverability measures the ability of a pruned model to recover accuracy. As shown in Fig.~\ref{fig:metric}, we will take the `layer1.1' block as an example. Simply dropping this block results in a 54.3\% Top-1 error. Both the KL-divergence (`CURL') and `L2 distance' variations before/after block removal are large, which indicates the model's outputs are indeed changed dramatically. However, the accuracy can be recovered effectively by finetuning with a tiny training set. These existing criteria cannot reveal this trend, which however directly determines compression quality. But ours (`PRACTISE') enjoys a high consistency with the Top-1 error of the finetuned model.

Fig.~\ref{fig:practise} presents our method on how to compute the recoverability. Given the original model $\mathcal{M}_O$, the pruned model $\mathcal{M}_{P(\mathcal{B}_i)}$ is obtained by dropping a block $\mathcal{B}_i$. To eliminate the effect of dropping this block with minimum efforts, we insert adaptors in the positions connected to this block. Surprisingly, we empirically find optimizing only adaptors is close to optimizing all parameters under the few-shot setting. Hence, our defined recoverability is calculated as
\begin{equation}
  \label{eq:recoverability}
  \mathcal{R}(\mathcal{B}_i) = \min_{\alpha} \mathbb{E}_{x\sim p(x)}\| \mathcal{M}_O(x;\theta) - \mathcal{M}_{P(\mathcal{B}_i)}(x;\theta\setminus b_i, \alpha)\|^2_F\,,
\end{equation}
where $\theta$ means parameters in the original model, and $\setminus b_i$ means excluding the parameters in the dropped block $\mathcal{B}_i$, and $\alpha$ denotes parameters in adaptors. All adaptors are conv. layers with kernel size $1\times 1$ and placed before/after the raw conv. layers according to different positions. For blocks in front of the dropped block, adaptors are inserted after conv. layers. On the contrary, adaptors are inserted before conv. layers. Because convolutions are linear, all adaptors can be fused in the neighbor conv. layers while keeping the outputs unchanged. That means these adaptors will \emph{not be overhead} to the pruned model. Thanks to the limited computations and parameters in adaptors, calculating the recoverability by Eq.~\ref{eq:recoverability} requires only a few training samples and little training time. 

\begin{figure}
	\centering
	\includegraphics[width=0.7\linewidth]{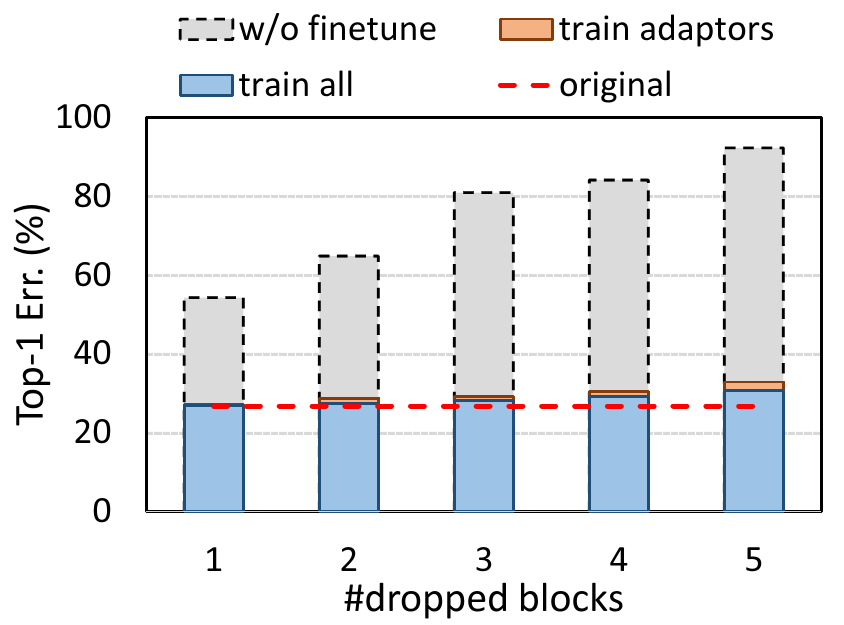}
	\caption{Comparison of training all and only adaptors. ResNet-34 was pruned by dropping different numbers of blocks with only 500 training samples. Top-1 error (\%) was evaluated on the ImageNet-1k validation set. Note that the Top-1 error of the original model is 26.98\%. }
	\label{fig:drop-5-blocks}
\end{figure}
	
Fig.~\ref{fig:metric} has demonstrated our method is effective to estimate the recoverability of each block. Fig.~\ref{fig:drop-5-blocks} shows these adaptors can recover most accuracy loss when dropping more blocks. Hence it results in an excellent metric to predict the accuracy of the finetuned model. But there is a principal problem. Because non-linear operations (\eg, ReLU) exist in blocks, in theory the linear adaptors cannot eliminate the dropping effect perfectly. So the empirical phenomenons suggest that \emph{finetuning the whole network on the tiny set mainly recovers the model's linearity part}, although CNNs are considered highly non-linear models. 

Another factor for dropping blocks we should take care of is the latencies of different blocks are different. With the same recoverability, the block with a higher latency should have a higher priority to drop. To this end, we calculate the acceleration ratio of the block $\mathcal{B}_i$ by 
\begin{equation}
  \label{eq:latency-ratio}
  \tau(\mathcal{B}_i) = \frac{lat_{\mathcal{M}_O} - lat_{\mathcal{M}_{P(\mathcal{B}_i)}}}{lat_{\mathcal{M}_O}}\,,
\end{equation}
where $lat$ denotes the latency. Finally, we define the pruning score for each block $\mathcal{B}_i$ as 
\begin{equation}
  \label{eq:score}
  s(\mathcal{B}_i) = \frac{\mathcal{R}(\mathcal{B}_i)}{\tau(\mathcal{B}_i)}\,.
\end{equation}
Our pruning score considers both the recoverability and latency of each block. A lower score means the block has a higher priority to drop. We compute the pruning score for each block and drop the top $k$ blocks with minimum scores.

\subsection{Recover the accuracy of the pruned model}

Once the pruned model's structure is determined, the last problem is how to recover the accuracy. A naive method is using the tiny training set to finetune with the cross-entropy loss. However, the pruned model easily suffers from overfitting, as previous works pointed out~\cite{FSKD,CD}. Knowledge distillation~\cite{KD}, especially feature-level distillation~\cite{fitnet,LSHKD}, can alleviate the overfitting problem and achieve superior accuracy. MiR~\cite{MiR} proposed to use the features before the global average pooling for the pruned model to mimic and achieved state-of-the-art performance with few training samples. We follow MiR and finetune the pruned model by minimizing
\begin{equation}
  \label{eq:ft-loss}
  \mathcal{L} = \| \mathcal{M}_O(x;\theta_O) - \mathcal{M}_P(x;\theta_P)\|^2_F\,,
\end{equation}
where $\theta_O$ denotes the frozen parameters of the original model and $\theta_P$ denotes the trainable parameters of the pruned model. 

\begin{algorithm}[t]
  \caption{\Practise}
  \label{alg:practise}
  \KwIn{The original model $\mathcal{M}_O$, the number of dropped blocks $k$, the tiny training data $\mathcal{D}_{\mathcal{T}}$}
  Test the latency of $\mathcal{M}_O$\;
  \For{each block $\mathcal{B}_i$}
  {
    Drop $\mathcal{B}_i$ to obtain the pruned model $\mathcal{M}_{P(\mathcal{B}_i)}$\;
    Test latency of $\mathcal{M}_{P(\mathcal{B}_i)}$ and find $\tau(\mathcal{B}_i)$ (Eq.~\ref{eq:latency-ratio})\;
    Insert adaptors\;
    Compute $\mathcal{R}(\mathcal{B}_i)$ with $\mathcal{D}_{\mathcal{T}}$ (Eq.~\ref{eq:recoverability})\;
    Compute the score $s(\mathcal{B}_i)$ (Eq.~\ref{eq:score})\;
    Add $\mathcal{B}_i$ back and remove all adaptors;
  }
  Choose the top $k$ blocks with the minimum scores\;
  Drop these $k$ blocks to obtain $\mathcal{M}_P$\;
  Finetune $\mathcal{M}_P$ with $\mathcal{D}_{\mathcal{T}}$ by minimizing $\mathcal{L}$ (Eq.~\ref{eq:ft-loss})\;
  \textbf{return} The pruned model $\mathcal{M}_P$
\end{algorithm}

Overall, the whole algorithm of \Practise is presented in Alg.~\ref{alg:practise}. Our \Practise enjoys \emph{zero} extra hyperparameters. With feature mimicking, we accelerate models \emph{without} using training labels.

\begin{table*}
	\centering
  \small
	\begin{tabular}{c|c|cccc}
		\toprule
		Method & Latency (ms) & 50	& 100 & 500	& 1000 \\
		\midrule
    BP (filter) & 35.1 (15.8\%$\downarrow$) & $39.0_{\pm 1.41}/68.9_{\pm 1.17}$ & $41.0_{\pm 0.33}/70.5_{\pm 0.66}$ & $51.8_{\pm 0.30}/78.1_{\pm 0.38}$ & $57.8_{\pm 0.30}/81.5_{\pm 0.18}$\\

    BP (block) & \textbf{34.9} (\textbf{16.3}\%$\downarrow$) & $\textbf{66.5}_{\pm 0.81}/\textbf{78.4}_{\pm 0.44}$ & $\textbf{66.8}_{\pm 0.23}/\textbf{87.7}_{\pm 0.23}$ & $\textbf{68.6}_{\pm 0.18}/\textbf{88.8}_{\pm 0.09}$ & $\textbf{69.8}_{\pm 0.12}/\textbf{89.3}_{\pm 0.07}$\\
    \midrule

    KD~\cite{KD} & 35.1 (15.8\%$\downarrow$) & $44.5_{\pm 1.20}/72.3_{\pm 0.87}$ & $46.4_{\pm 0.34}/74.0_{\pm 0.58}$ & $54.7_{\pm 0.26}/79.7_{\pm 0.19}$ & $57.9_{\pm 0.21}/81.6_{\pm 0.12}$\\

    FSKD~\cite{FSKD} & 35.1 (15.8\%$\downarrow$) & $45.3_{\pm 0.77}/71.5_{\pm 0.62}$ & $51.2_{\pm 0.30}/76.8_{\pm 0.23}$ & $57.6_{\pm 0.21}/81.6_{\pm 0.15}$ & $59.4_{\pm 0.13}/82.7_{\pm 0.06}$\\

    CD~\cite{CD} & 35.1 (15.8\%$\downarrow$) & $56.2_{\pm 0.37}/80.8_{\pm 0.31}$ & $59.1_{\pm 0.22}/82.8_{\pm 0.11}$ & $63.7_{\pm 0.18}/86.0_{\pm 0.05}$ & $64.4_{\pm 0.03}/86.3_{\pm 0.07}$\\

    MiR~\cite{MiR} & 35.1 (15.8\%$\downarrow$) & $64.1_{\pm 0.10}/86.3_{\pm 0.11}$ & $65.1_{\pm 0.19}/87.0_{\pm 0.11}$ & $67.0_{\pm 0.09}/88.1_{\pm 0.07}$ & $67.8_{\pm 0.06}/88.5_{\pm 0.02}$\\

    \Practise & \textbf{34.9} (\textbf{16.3}\%$\downarrow$) & $\textbf{70.3}_{\pm 0.16}/\textbf{89.6}_{\pm 0.06}$ & $\textbf{71.5}_{\pm 0.74}/\textbf{90.3}_{\pm 0.37}$ & $\textbf{72.5}_{\pm 0.04}/\textbf{90.9}_{\pm 0.03}$ & $\textbf{72.5}_{\pm 0.05}/\textbf{91.0}_{\pm 0.02}$ \\
		\bottomrule
	\end{tabular}
  \caption{Top-1/Top-5 validation accuracy (\%) on ImageNet-1k for pruning ResNet-34. All models were accelerated by reducing about $16\%$ latency with 50, 100, 500, and 1000 training samples (in the top row). Previous methods pruned filters within the residual block (\ie, `normal' in Fig.~\ref{fig:latency-flops}). The Top-1/Top-5 accuracy of the original ResNet-34 are $73.31\%/91.42\%$. }
  \label{tab:ImageNet-ResNet34-highlan-4data}
\end{table*}

\begin{table*}
	\centering
  \small
	\begin{tabular}{c|c|cccc}
		\toprule
		Method & Latency (ms) & 50	& 100 & 500	& 1000 \\
		\midrule
    BP (filter) & $33.8\,(18.9\%\downarrow)$ & $24.2_{\pm 0.92}/52.7_{\pm 1.36}$ & $27.6_{\pm 0.41}/56.7_{\pm 0.62}$ & $42.9_{\pm 0.28}/70.5_{\pm 0.27}$ & $51.2_{\pm 0.32}/76.5_{\pm 0.16}$\\

    BP (block) & $\textbf{32.5}\,(\textbf{22.1}\%\downarrow)$ & $\textbf{60.6}_{\pm 0.62}/\textbf{83.5}_{\pm 0.42}$ & $\textbf{61.6}_{\pm 0.31}/\textbf{84.3}_{\pm 0.36}$ & $\textbf{65.0}_{\pm 0.19}/\textbf{86.5}_{\pm 0.20}$ & $\textbf{66.8}_{\pm 0.18}/\textbf{87.5}_{\pm 0.13}$\\
    \midrule

    KD~\cite{KD} & $33.8\,(18.9\%\downarrow)$ & $30.1_{\pm 0.69}/57.7_{\pm 1.10}$ & $33.1_{\pm 0.43}/61.0_{\pm 0.53}$ & $45.7_{\pm 0.26}/72.2_{\pm 0.25}$ & $50.5_{\pm 0.29}/75.9_{\pm 0.23}$\\

    FSKD~\cite{FSKD} & $33.8\,(18.9\%\downarrow)$ & $31.1_{\pm 0.90}/56.5_{\pm 1.10}$ & $36.6_{\pm 0.44}/63.1_{\pm 0.46}$ & $42.8_{\pm 0.49}/69.1_{\pm 0.58}$ & $44.9_{\pm 0.20}/70.5_{\pm 0.29}$\\

    MiR~\cite{MiR} & $33.8\,(18.9\%\downarrow)$ & $59.9_{\pm 0.30}/83.2_{\pm 0.31}$ & $62.1_{\pm 0.22}/84.8_{\pm 0.18}$ & $65.4_{\pm 0.07}/87.0_{\pm 0.03}$ & $66.6_{\pm 0.05}/87.7_{\pm 0.04}$\\

    \Practise & $\textbf{32.5}\,(\textbf{22.1}\%\downarrow)$ & $\textbf{68.0}_{\pm 1.36}/\textbf{88.2}_{\pm 0.77}$ & $\textbf{70.4}_{\pm 0.42}/\textbf{89.7}_{\pm 0.23}$ & $\textbf{71.8}_{\pm 0.07}/\textbf{90.5}_{\pm 0.02}$ & $\textbf{71.9}_{\pm 0.05}/\textbf{90.6}_{\pm 0.04}$ \\
		\bottomrule
	\end{tabular}
  \caption{Top-1/Top-5 validation accuracy (\%) on ImageNet-1k for pruning ResNet-34. Our model was accelerated by reducing $22.1\%$ latency. Previous methods pruned filters both inside and outside the residual connection (\ie, `residual' in Fig.~\ref{fig:latency-flops}). The Top-1/Top-5 accuracy of the original ResNet-34 are $73.31\%/91.42\%$. }
  \label{tab:ImageNet-ResNet34-lowlan-4data}
\end{table*}

Our \Practise can even work in data-free scenarios. One choice is treating the synthesized images from DFKD methods~\cite{DI} as the training images. Most existing works~\cite{DI,MixMix} adopt filter pruning, whereas our \Practise improves the latency-accuracy performance by a significant margin. That helps a lot in data-free scenarios. Another choice is collecting out-of-domain data. On the other hand, with a large amount of out-of-domain images, the accuracy of the pruned network is even close to that of using original training images. That demonstrates the high generalization ability of \PractiseNoSpace.

\section{Experimental Results}
\label{sec:exp}

\begin{table*}
	\centering
  \small
	\begin{tabular}{c|r@{.}lr@{.}lr@{.}lr@{.}lr@{.}lr@{.}l}
		\toprule
		Latency (ms) & \multicolumn{2}{c}{BP (filter)} & \multicolumn{2}{c}{KD~\cite{KD}} & \multicolumn{2}{c}{FSKD~\cite{FSKD}}	& \multicolumn{2}{c}{CD~\cite{CD}} & \multicolumn{2}{c}{MiR~\cite{MiR}} & \multicolumn{2}{c}{\Practise}	\\
		\midrule
    32.5 (22.1\%$\downarrow$) & $47$&$54_{\pm 0.41}$ & $49$&$34_{\pm 0.25}$ & $33$&$19_{\pm 0.60}$ & $59$&$65_{\pm 0.12}$ & $68$&$14_{\pm 0.04}$ & $\mathbf{71}$&$\mathbf{75}_{\pm 0.07}$ \\

    34.9 (16.3\%$\downarrow$) & $58$&$94_{\pm 0.36}$ & $61$&$01_{\pm 0.24}$ & $62$&$56_{\pm 0.13}$ & $68$&$17_{\pm 0.07}$ & $70$&$53_{\pm 0.10}$ & $\mathbf{72}$&$\mathbf{52}_{\pm 0.04}$ \\

    38.3 \,\,\,(8.2\%$\downarrow$) & $65$&$02_{\pm 0.30}$ & $67$&$22_{\pm 0.18}$ & $69$&$59_{\pm 0.09}$ & $71$&$12_{\pm 0.06}$ & $71$&$95_{\pm 0.07}$ & $\mathbf{73}$&$\mathbf{04}_{\pm 0.06}$ \\

		\bottomrule
	\end{tabular}
  \caption{Top-1 validation accuracy (\%) on ImageNet-1k for pruning ResNet-34. The model was pruned at three different compression ratios with 500 training samples. Previous methods prune filters by CD-style, while we drop blocks. The Top-1 accuracy and the latency of the original ResNet-34 are $73.31\%$ and $41.7$ ms, respectively. }
  \label{tab:ImageNet-ResNet34-Top1-LA-500}
\end{table*}

In this section, we evaluate the performance of \PractiseNoSpace. Following previous works~\cite{FSKD,CD,MiR}, ResNet-34~\cite{ResNet} and MobileNetV2~\cite{mobilenet} will be pruned on tiny training sets of ImageNet-1k~\cite{imagenet}. Then, to test the generalization ability of \PractiseNoSpace, we will prune ResNet-50 with synthesized images and out-of-domain data, respectively. Finally, ablation studies are conducted for further analyzing \PractiseNoSpace.

\textbf{Implementation details.} As presented in Alg.~\ref{alg:practise}, \Practise requires computing the latency and recoverability of each pruned model $\mathcal{M}_{P(\mathcal{B}_i)}$, then finetuning the pruned model $\mathcal{M}_{P}$. In this paper, we only consider dropping the block with the same input and output dimensionality. For the latency, we tested the model with the $64\times 3\times 224\times 224$ input by 500 times and take the mean as the latency number. Note that all latency numbers in this paper were tested on the same computer with an NVIDIA TITAN Xp GPU. To compute the recoverability in Eq.~\ref{eq:recoverability}, we used SGD with batch size $64$ to optimize adaptors by 1000 iterations. The initial learning rate was $0.02$ and decreased by a factor of 10 per 40\% iterations. Note that if the size of training data $\mathcal{D}_{\mathcal{T}}$ is less than $64$, the batch size will equal this number. After optimizing adaptors, $\mathcal{R}(\mathcal{B}_i)$ was computed by Eq.~\ref{eq:recoverability} with the training set $\mathcal{D}_{\mathcal{T}}$. For the final finetuning, all parameters in the pruned network were updated by SGD with minimizing $\mathcal{L}$ in Eq.~\ref{eq:ft-loss}. Finetuning ran 2000 iterations by default. The settings of batch size and learning rate schedules were the same as those of optimizing adaptors. For a fair comparison, we used the data argumentation strategy supplied by PyTorch official examples, which is the same as that of previous works~\cite{FSKD,CD,MiR}. All experiments were conducted with PyTorch~\cite{pytorch}.

To compare with other few-shot pruning methods, we are mainly concerned about the latency-accuracy tradeoff of the pruned model. We tested the latency of networks pruned by previous methods and directly cite their accuracy number. The results of \Practise were run by five times with different sampled tiny training sets. We report the mean accuracy along with standard deviation. 

\subsection{Different amounts of training data}

We compare our \Practise with the state-of-the-art few-shot pruning methods. Following the previous setting, we prune ResNet-34 with different amounts of training images on ImageNet-1k. Table~\ref{tab:ImageNet-ResNet34-highlan-4data} summarizes results. All previous methods pruned filters within residual blocks (cf. `normal' in Fig.~\ref{fig:latency-flops}), whereas our dropping blocks achieves better results. First, we just removed the first few blocks to reach the latency goal and then simply finetuned the network with the cross-entropy loss. This simple baseline for dropping blocks, `BP (block)', has already outperformed previous methods. Note that this is our first contribution that revealing the advantage of dropping blocks in the few-shot scenario. \Practise further improves results. With similar latency reduction, it dramatically outperforms previous SOTA by an average of $5.7\%$ Top-1 accuracy on ImageNet-1k. Table~\ref{tab:ImageNet-ResNet34-lowlan-4data} compares these methods with a larger latency reduction. \Practise surpasses MiR by a significant margin again, on average $7.0\%$ Top-1 accuracy. Note that our model even is faster than previous ones by $1.3$ ms. Both Table~\ref{tab:ImageNet-ResNet34-highlan-4data} and \ref{tab:ImageNet-ResNet34-lowlan-4data} demonstrate that dropping blocks is a superior manner for accelerating networks with tiny sets. 

\subsection{Different acceleration ratios}

Table~\ref{tab:ImageNet-ResNet34-Top1-LA-500} compares \Practise with previous methods for different acceleration ratios. For $8.2\%$ latency reduction, \Practise outperforms MiR by $1.1\%$ Top-1 accuracy. In further reducing latency by $22.1\%$, \Practise surpasses MiR by $3.6\%$ Top-1 accuracy. That indicates \Practise enjoys higher accuracy than others when the acceleration ratio becomes larger. Fig.~\ref{fig:top1-latency} presents curves for the latency-accuracy tradeoffs of different methods. Our \Practise outperforms previous methods dominantly and makes a new milestone in the field of few-shot compression. 

\subsection{The data-latency-accuracy tradeoff}

\begin{figure}
	\centering
	\includegraphics[width=0.8\linewidth]{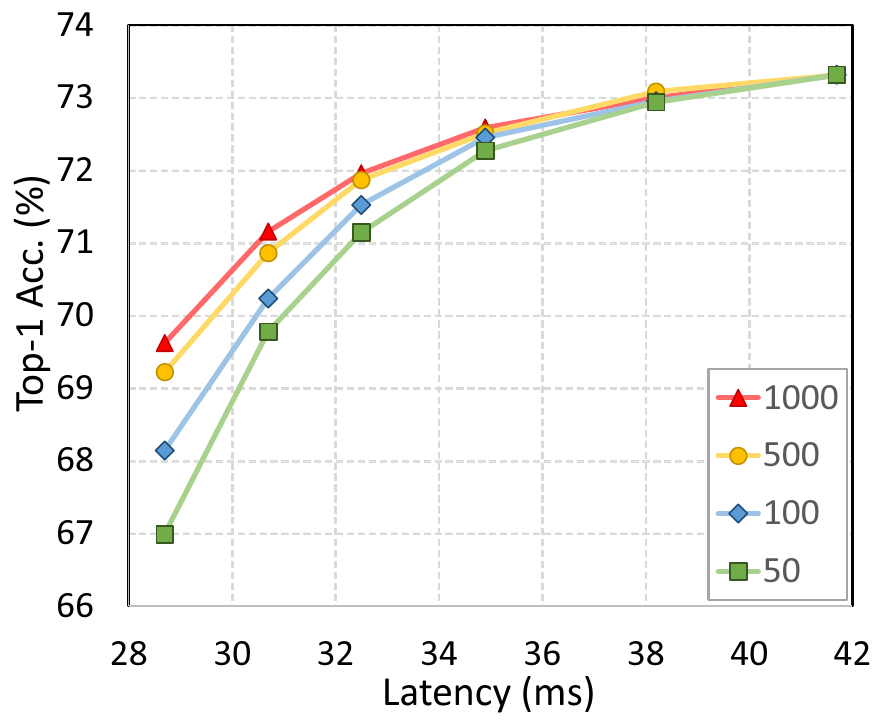}
	\caption{Illustration of the data-latency-accuracy tradeoff. ResNet-34 was pruned by \Practise on ImageNet-1k. Numbers in the legend mean different amounts of training images. }
	\label{fig:data_latency_acc}
  \vspace{-1mm}
\end{figure}

Previous experiments have demonstrated the advantage of \Practise. Next, for a better understanding of the data-latency-accuracy tradeoff in the few-shot compression scenario, we pruned ResNet-34 by \Practise with different latency reductions and amounts of training images. Fig.~\ref{fig:data_latency_acc} presents the results. With less latency reduction, the accuracies for different amounts of training data are comparable. But for a large latency reduction, it is challenging to recover the accuracy by only a tiny training set. And the accuracy gap becomes large w.r.t. different amounts of training data.  

\subsection{Results on MobileNetV2}

\begin{table}
	\centering
  \small
	\begin{tabular}{c|cc}
		\toprule
		Method & Latency (ms) & Top-1/Top-5	\\
		\midrule
    Original & \textcolor{lightgray}{37.6} & \textcolor{lightgray}{71.9/90.3} \\
    \midrule
    BP (filter) & $31.5\,(16.2\%\downarrow)$ & $45.0_{\pm 0.34}/71.8_{\pm 0.38}$ \\

    KD~\cite{KD} & $31.5\,(16.2\%\downarrow)$ & $48.4_{\pm 0.34}/73.9_{\pm 0.32}$ \\

    MiR~\cite{MiR} & $31.5\,(16.2\%\downarrow)$ & $67.6_{\pm 0.05}/87.9_{\pm 0.04}$ \\

    \Practise & $\textbf{30.4}\,(\textbf{19.1}\%\downarrow)$ & $\textbf{69.3}_{\pm 0.05}/\textbf{88.9}_{\pm 0.05}$ \\
    \midrule
    BP (filter) & $34.1\,\,\,(9.3\%\downarrow)$ & $55.5_{\pm 0.16}/80.3_{\pm 0.26}$ \\

    KD~\cite{KD} & $34.1\,\,\,(9.3\%\downarrow)$ & $59.1_{\pm 0.17}/82.5_{\pm 0.15}$ \\

    MiR~\cite{MiR} & $34.1\,\,\,(9.3\%\downarrow)$ & $69.7_{\pm 0.04}/89.2_{\pm 0.03}$ \\

    \Practise & $\textbf{31.9}\,(\textbf{15.2}\%\downarrow)$ & $\textbf{70.3}_{\pm 0.03}/\textbf{89.5}_{\pm 0.03}$ \\
		\bottomrule
	\end{tabular}
  \caption{Comparison of \Practise and few-shot pruning methods on ImageNet-1k. MobileNetV2 was pruned with 500 samples. }
  \label{tab:ImageNet-MobileNetV2}
  \vspace{-1mm}
\end{table}

MobileNetV2~\cite{mobilenet} is a lightweight model and is popularly applied on mobile devices. It has 10 blocks which can be dropped by our \PractiseNoSpace. Table~\ref{tab:ImageNet-MobileNetV2} summarizes results. \Practise outperforms previous methods by a significant margin. Compared with MiR, \Practise obtains pruned models with both lower latency and higher accuracy. 

\subsection{Train with synthesized/out-of-domain images}

\begin{table}[!htb]
	\centering
  \setlength{\tabcolsep}{2pt}
  \small
	\begin{tabular}{c|cccc}
		\toprule
		Network & Method & Pruning & Latency & Top-1	\\
		\midrule
    \multirow{6}*{ResNet-50} & 
    Original & & \textcolor{lightgray}{83.8} & \textcolor{lightgray}{76.1} \\
    & DI~\cite{DI} & filter & - & $72.0$ \\
    & MixMix~\cite{MixMix} & filter & - & $69.8$ \\
    & ADI~\cite{DI} & filter & - & $73.3$ \\
    & ADI*~\cite{DI} & filter & 79.9 \,\,\,(4.7\%$\downarrow$) & $73.5$ \\
    & \Practise & block & \textbf{66.2} (\textbf{21.0}\%$\downarrow$) & \textbf{74.8} \\
    \midrule
    \multirow{5}*{MobileNetV2} & 
    Original & & \textcolor{lightgray}{37.6} & \textcolor{lightgray}{71.9} \\ 
    & DI~\cite{DI} & filter & - & 15.3 \\
    & MixMix~\cite{MixMix} & filter & - & 42.5 \\
    & ADI*~\cite{DI} & filter & 30.8 (18.1\%$\downarrow$) & 62.8 \\
    & \Practise & block & \textbf{30.4} (\textbf{19.1}\%$\downarrow$) & \textbf{68.0} \\
		\bottomrule
	\end{tabular}
  \caption{Comparison of \Practise and data-free methods on ImageNet-1k. Latency (ms) and Top-1 validation accuracy (\%) on ImageNet-1k are reported. * denotes the method reimplemented by ourselves. }
  \label{tab:ADI}
\end{table}

\begin{table}[!htb]
	\centering
  \small
	\begin{tabular}{c|ccccc}
		\toprule
		Dateset & 50 & 500 & 1000 & 5000 & All \\
		\midrule
    ImageNet~\cite{imagenet} & \textcolor{lightgray}{74.22} & \textcolor{lightgray}{74.58} & \textcolor{lightgray}{74.58} & \textcolor{lightgray}{75.14} & \textcolor{lightgray}{75.24} \\
    ADI~\cite{DI}  & 69.85 & 72.68 & 73.01 & 74.40 & 74.79 \\
    CUB~\cite{CUB} & 72.49 & 73.71 & 73.94 & 74.86 & 74.92 \\
    Place365~\cite{places365} & \textbf{72.80} & \textbf{74.10} & \textbf{74.18} & \textbf{75.05} & \textbf{75.21} \\
		\bottomrule
	\end{tabular}
  \caption{Top-1 validation accuracy (\%) on ImageNet-1k for different out-of-domain training datasets. ResNet-50 was accelerated by reducing $21\%$ latency with \PractiseNoSpace. }
  \label{tab:Otherdata-ResNet50}
\end{table}

In some extreme scenarios, not even a single original training sample is to be provided. Zero-shot pruning is required. Because \Practise does not need ground-truth labels, it is able to accelerate networks with the images synthesized by data-free knowledge distillation methods. Table~\ref{tab:ADI} summarizes results. Note that most zero-shot pruning methods adopt filter pruning, which are inferior to pruning blocks as we have shown. We adopt synthesized images produced by ADI~\cite{DI} as the training set. For both ResNet-50 and MobileNetV2, our \Practise achieves higher Top-1 accuracy with more latency reductions. And we advocate that we should adopt dropping blocks for DFKD in the future to accelerate networks more effectively.

Another choice of zero-shot pruning is collecting out-of-domain training images. Our \Practise is also robust to work with these data. Table~\ref{tab:Otherdata-ResNet50} presents results. The original ResNet-50 was trained on ImageNet-1k, and we pruned it on other datasets by \PractiseNoSpace. ADI consists of images synthesized by DeepInversion~\cite{DI}. CUB~\cite{CUB} contains images of birds with 200 categories. Place365~\cite{places365} consists of scene pictures. Our \Practise enjoys a high generalization ability to work with all these datasets. Another benefit of the out-of-domain data is the unlimited number of images. We notice the accuracy is boosted by using more training samples and even close to that of using original training data.

\subsection{Different criteria for dropping blocks}

\begin{table}
	\centering
  \small
	\begin{tabular}{c|ccccc}
		\toprule
		\#Dropped blocks & 1 & 2 & 3 & 4	& 5 \\
		\midrule
    Random  & 71.12 & 71.27 & 67.89 & 64.27 & 63.47 \\
    CURL~\cite{CURL}  & 72.33 & 71.08 & 69.14 & 65.48 & 64.97 \\
    $\epsilon$-ResNet~\cite{e_ResNet} & 72.51 & 71.20 & 69.00 & 67.90 & 64.75 \\
    L2 distance       & 72.51 & 71.20 & 69.00 & 68.61 & 64.93 \\
    \Practise         & \textbf{73.02} & \textbf{72.52} & \textbf{71.92} & \textbf{70.86} & \textbf{69.35} \\
		\bottomrule
	\end{tabular}
  \caption{Top-1 validation accuracy (\%) on ImageNet-1k for pruning ResNet-34 with 500 images. We compare different criteria for dropping different numbers of blocks. }
  \label{tab:ImageNet-ResNet34-choose-block}
\end{table}

Finally, we compare \Practise with other criteria for dropping blocks. Fig.~\ref{fig:metric} shows results for removing only one block. Obviously, \Practise is better than others, and is very consistent with the finetuned models' accuracies. Most existing methods mainly measure the gap between the original network and the pruned network without finetuning. They neglect the recoverability of each dropped block, hence resulting in an inferior strategy for dropping blocks. Our \Practise pays attention to the recoverability and the acceleration ratio of each block. Therefore the pruned network enjoys higher accuracy and lower latency. 

Table~\ref{tab:ImageNet-ResNet34-choose-block} compares different criteria for dropping more blocks. As the number of dropped blocks increases, our \Practise outperforms others by even larger margins. To sum up, our \Practise is able to find inefficient blocks to drop, compared with other methods.

\section{Conclusions and Future Works}

This paper aims at accelerating networks with tiny training sets. For the first time we revealed that dropping blocks is more effective than previous filter-level pruning in this scenario. We believe this finding makes significant progress in the few-shot model compression. To determine which blocks to drop, we proposed a new concept namely \emph{recoverability} to measure the difficulty of recovering the pruned network's accuracy with few samples. Compared with previous pruning criteria, our recoverability is more related to the model's accuracy after finetuning. Our method for computing recoverability reveals that the end-to-end finetuning with a tiny set mainly recovers the model's linear ability. Finally, \Practise was proposed. It enjoys high latency-accuracy performance and is robust to deal with synthesized/out-of-domain images. Extensive experiments demonstrated that \Practise outperforms previous methods by a significant margin (on average 7\% Top-1 accuracy on ImageNet-1k for 22\% latency reduction).

\Practise has limitations such as it is confined to recognition, which leads to future explorations. It is promising to extend \Practise for other models (\eg, Transformer) or other vision tasks (\eg, object detection and segmentation). For network acceleration, how to compute the recoverability of other compression schemes is also an interesting problem. Recently, finetuning and accelerating a large pretrained network on downstream tasks are emerging as critical needs. 
Applying \Practise for tuning a pretrained model on downstream tasks is also promising.

{\small
\bibliographystyle{ieee_fullname}
\bibliography{egbib}

\begin{thebibliography}{10}\itemsep=-1pt

\bibitem{CD}
Haoli Bai, Jiaxiang Wu, Irwin King, and Michael Lyu.
\newblock Few shot network compression via cross distillation.
\newblock In {\em AAAI}, volume~04, pages 3203--3210, 2020.

\bibitem{bannerscalable}
Ron Banner, Itay Hubara, Elad Hoffer, and Daniel Soudry.
\newblock Scalable methods for 8-bit training of neural networks.
\newblock In {\em NeurIPS\ 31}, pages 5151--5159, 2018.

\bibitem{cai2020zeroq}
Yaohui Cai, Zhewei Yao, Zhen Dong, Amir Gholami, Michael~W Mahoney, and Kurt
  Keutzer.
\newblock {ZeroQ}: A novel zero shot quantization framework.
\newblock In {\em CVPR}, pages 13169--13178, 2020.

\bibitem{ResRep}
Xiaohan Ding, Tianxiang Hao, Jianchao Tan, Ji Liu, Jungong Han, Yuchen Guo, and
  Guiguang Ding.
\newblock {ResRep}: Lossless cnn pruning via decoupling remembering and
  forgetting.
\newblock In {\em ICCV}, pages 4510--4520, 2021.

\bibitem{fang2022up}
Gongfan Fang, Kanya Mo, Xinchao Wang, Jie Song, Shitao Bei, Haofei Zhang, and
  Mingli Song.
\newblock Up to 100x faster data-free knowledge distillation.
\newblock In {\em AAAI}, volume~6, pages 6597--6604, 2022.

\bibitem{DW_PW}
Jianbo Guo, Yuxi Li, Weiyao Lin, Yurong Chen, and Jianguo Li.
\newblock Network decoupling: From regular to depthwise separable convolutions.
\newblock In {\em BMVC}, page 248, 2018.

\bibitem{quantization}
Song Han, Huizi Mao, and William~J Dally.
\newblock Deep compression: Compressing deep neural networks with pruning,
  trained quantization and huffman coding.
\newblock In {\em ICLR}, pages 1--14, 2016.

\bibitem{ResNet}
Kaiming He, Xiangyu Zhang, Shaoqing Ren, and Jian Sun.
\newblock Deep residual learning for image recognition.
\newblock In {\em CVPR}, pages 770--778, 2016.

\bibitem{he2019filter}
Yang He, Ping Liu, Ziwei Wang, Zhilan Hu, and Yi Yang.
\newblock Filter pruning via geometric median for deep convolutional neural
  networks acceleration.
\newblock In {\em CVPR}, pages 4340--4349, 2019.

\bibitem{KD}
Geoffrey~E. Hinton, Oriol Vinyals, and Jeffrey Dean.
\newblock Distilling the knowledge in a neural network.
\newblock {\em arXiv preprint arXiv:1503.02531}, 2015.

\bibitem{L1_norm}
Hao Li, Asim Kadav, Igor Durdanovic, Hanan Samet, and Hans~Peter Graf.
\newblock Pruning filters for efficient convnets.
\newblock {\em arXiv preprint arXiv:1608.08710}, 2016.

\bibitem{FSKD}
Tianhong Li, Jianguo Li, Zhuang Liu, and Changshui Zhang.
\newblock Few sample knowledge distillation for efficient network compression.
\newblock In {\em CVPR}, pages 14639--14647, 2020.

\bibitem{li2022revisiting}
Yawei Li, Kamil Adamczewski, Wen Li, Shuhang Gu, Radu Timofte, and Luc
  Van~Gool.
\newblock Revisiting random channel pruning for neural network compression.
\newblock In {\em CVPR}, pages 191--201, 2022.

\bibitem{MixMix}
Yuhang Li, Feng Zhu, Ruihao Gong, Mingzhu Shen, Xin Dong, Fengwei Yu, Shaoqing
  Lu, and Shi Gu.
\newblock {MixMix}: All you need for data-free compression are feature and data
  mixing.
\newblock In {\em ICCV}, pages 4410--4419, 2021.

\bibitem{LRD}
Shaohui Lin, Rongrong Ji, Chao Chen, Dacheng Tao, and Jiebo Luo.
\newblock Holistic {CNN} compression via low-rank decomposition with knowledge
  transfer.
\newblock {\em IEEE TPAMI}, 41(12):2889--2905, 2018.

\bibitem{liu2021discrimination}
Jing Liu, Bohan Zhuang, Zhuangwei Zhuang, Yong Guo, Junzhou Huang, Jinhui Zhu,
  and Mingkui Tan.
\newblock Discrimination-aware network pruning for deep model compression.
\newblock {\em IEEE TPAMI}, 44(8):4035--4051, 2021.

\bibitem{liu2021zero}
Yuang Liu, Wei Zhang, and Jun Wang.
\newblock Zero-shot adversarial quantization.
\newblock In {\em CVPR}, pages 1512--1521, 2021.

\bibitem{slimming}
Zhuang Liu, Jianguo Li, Zhiqiang Shen, Gao Huang, Shoumeng Yan, and Changshui
  Zhang.
\newblock Learning efficient convolutional networks through network slimming.
\newblock In {\em ICCV}, pages 2736--2744, 2017.

\bibitem{liu2018rethinking}
Zhuang Liu, Mingjie Sun, Tinghui Zhou, Gao Huang, and Trevor Darrell.
\newblock Rethinking the value of network pruning.
\newblock In {\em ICLR}, pages 1--21, 2018.

\bibitem{JointPruning}
Zechun Liu, Xiangyu Zhang, Zhiqiang Shen, Yichen Wei, Kwang-Ting Cheng, and
  Jian Sun.
\newblock Joint multi-dimension pruning via numerical gradient update.
\newblock {\em IEEE TIP}, 30:8034--8045, 2021.

\bibitem{CURL}
Jian-Hao Luo and Jianxin Wu.
\newblock Neural network pruning with residual-connections and limited-data.
\newblock In {\em CVPR}, pages 1458--1467, 2020.

\bibitem{thinet}
Jian-Hao Luo, Jianxin Wu, and Weiyao Lin.
\newblock {ThiNet}: A filter level pruning method for deep neural network
  compression.
\newblock In {\em ICCV}, pages 5058--5066, 2017.

\bibitem{molchanov2019importance}
Pavlo Molchanov, Arun Mallya, Stephen Tyree, Iuri Frosio, and Jan Kautz.
\newblock Importance estimation for neural network pruning.
\newblock In {\em CVPR}, pages 11264--11272, 2019.

\bibitem{pytorch}
Adam Paszke, Sam Gross, Francisco Massa, Adam Lerer, James Bradbury, Gregory
  Chanan, Trevor Killeen, Zeming Lin, Natalia Gimelshein, Luca Antiga, et~al.
\newblock {PyTorch}: An imperative style, high-performance deep learning
  library.
\newblock In {\em NeurIPS\ 32}, pages 8026--8037, 2019.

\bibitem{fitnet}
Adriana Romero, Nicolas Ballas, Samira~Ebrahimi Kahou, Antoine Chassang, Carlo
  Gatta, and Yoshua Bengio.
\newblock {FitNets}: Hints for thin deep nets.
\newblock In {\em ICLR}, pages 1--13, 2015.

\bibitem{imagenet}
Olga Russakovsky, Jia Deng, Hao Su, Jonathan Krause, Sanjeev Satheesh, Sean Ma,
  Zhiheng Huang, Andrej Karpathy, Aditya Khosla, Michael Bernstein,
  Alexander~C. Berg, and Li Fei-Fei.
\newblock {ImageNet} large scale visual recognition challenge.
\newblock {\em IJCV}, 115(3):211--252, 2015.

\bibitem{mobilenet}
Mark Sandler, Andrew Howard, Menglong Zhu, Andrey Zhmoginov, and Liang-Chieh
  Chen.
\newblock {MobileNetV2}: Inverted residuals and linear bottlenecks.
\newblock In {\em CVPR}, pages 4510--4520, 2018.

\bibitem{CUB}
Catherine Wah, Steve Branson, Peter Welinder, Pietro Perona, and Serge
  Belongie.
\newblock The {Caltech-UCSD} birds-200-2011 dataset.
\newblock Technical Report CNS-TR-2011-001, California Institute of Technology,
  2011.

\bibitem{LSHKD}
Guo-Hua Wang, Yifan Ge, and Jianxin Wu.
\newblock Distilling knowledge by mimicking features.
\newblock {\em IEEE TPAMI}, 44(11):8183--8195, 2022.

\bibitem{MiR}
Huanyu Wang, Junjie Liu, Xin Ma, Yang Yong, Zhenhua Chai, and Jianxin Wu.
\newblock Compressing models with few samples: Mimicking then replacing.
\newblock In {\em CVPR}, pages 701--710, 2022.

\bibitem{DBP}
Wenxiao Wang, Shuai Zhao, Minghao Chen, Jinming Hu, Deng Cai, and Haifeng Liu.
\newblock {DBP}: Discrimination based block-level pruning for deep model
  acceleration.
\newblock {\em arXiv preprint arXiv:1912.10178}, 2019.

\bibitem{wu2018blockdrop}
Zuxuan Wu, Tushar Nagarajan, Abhishek Kumar, Steven Rennie, Larry~S Davis,
  Kristen Grauman, and Rogerio Feris.
\newblock Blockdrop: Dynamic inference paths in residual networks.
\newblock In {\em CVPR}, pages 8817--8826, 2018.

\bibitem{DI}
Hongxu Yin, Pavlo Molchanov, Jose~M Alvarez, Zhizhong Li, Arun Mallya, Derek
  Hoiem, Niraj~K Jha, and Jan Kautz.
\newblock Dreaming to distill: Data-free knowledge transfer via
  {DeepInversion}.
\newblock In {\em CVPR}, pages 8715--8724, 2020.

\bibitem{e_ResNet}
Xin Yu, Zhiding Yu, and Srikumar Ramalingam.
\newblock Learning strict identity mappings in deep residual networks.
\newblock In {\em CVPR}, pages 4432--4440, 2018.

\bibitem{zhang2021diversifying}
Xiangguo Zhang, Haotong Qin, Yifu Ding, Ruihao Gong, Qinghua Yan, Renshuai Tao,
  Yuhang Li, Fengwei Yu, and Xianglong Liu.
\newblock Diversifying sample generation for accurate data-free quantization.
\newblock In {\em CVPR}, pages 15658--15667, 2021.

\bibitem{places365}
Bolei Zhou, Aditya Khosla, Agata Lapedriza, Antonio Torralba, and Aude Oliva.
\newblock Places: An image database for deep scene understanding.
\newblock {\em arXiv preprint arXiv:1610.02055}, 2016.

\end{thebibliography}
}

\appendix

\clearpage
\section{Data Sampling}

Following previous work, our tiny training set is sampled uniformly from the whole set. In Tab.~\ref{tab:data-sample}, ``1-way N-shot'' denotes choosing one class randomly and then sampling N images from this class---the tiny set contains patterns of one specific class. It is indeed worse than uniform sampling, but the results are still acceptable, which means \Practise is robust with data from limited classes.

\begin{table}[!h]
  \centering
    \small
    \begin{tabular}{c|ccccc}
      \toprule
      Data sampling & N=100 & 500 & 1000 \\
      \midrule
      1-way N-shot & $69.8_{\pm 0.47}$ & $70.9_{\pm 0.42}$ & $70.0_{\pm 1.84}$ \\
      uniform & $\textbf{70.4}_{\pm 0.42}$ & $\textbf{71.8}_{\pm 0.07}$ & $\textbf{71.9}_{\pm 0.05}$ \\
    \bottomrule
    \end{tabular}
    \caption{Comparisons for different data sampling strategies. }
    \label{tab:data-sample}
\end{table}

\section{Training Time}

Evaluating the latency is efficient. The latency of raw ResNet-34 is 42ms, and testing it by 500 times costs only 21 seconds. Evaluating all 12 blocks requires about 5 minutes. Because of the tiny training set and limited training iterations, optimizing the model is also fast. Tab.~\ref{tab:train-time} reports the costed time on one Titan Xp GPU. Computing one block's recoverability only takes about 6 minutes. The total training time of \Practise is only about 1.5 hours.

\begin{table}[h]
  \centering
    \small
     \begin{tabular}{ccc|c}
     \toprule
     Latency & Recoverability & Finetuning & Total \\
     \midrule
     $0.4\times 12$ & $5.8\times 12$  & 11.3 & 85.7 \\
     \bottomrule
     \end{tabular}
  \caption{Training time (min). }
  \label{tab:train-time}
\end{table}

\section{Different Training Settings}

Here, we conduct ablation studies for the learning rate $\gamma$ and iterations. Tab.~\ref{tab:hyperparameters} presents results. We find that optimizing adaptors with 100 iterations is good enough while training all parameters requires more iterations.

\begin{table}[!htb]
   \centering
     \small
      \begin{tabular}{cc|ccc}
      \toprule
      Opt Eq.1 & Opt Eq.4 & $\gamma=0.01$ & 0.02 & 0.04 \\
      \midrule
      100 & 1000 & 71.14 & 71.65 & 71.26 \\
      100 & 2000 & 71.52 & \textbf{71.83} & 71.46 \\
      1000 & 2000 & 71.61 & 71.82 & 71.37 \\
      \bottomrule
      \end{tabular}
   \caption{Ablative results of different hyperparameters. }
   \label{tab:hyperparameters}
\end{table}

\end{document}